\RequirePackage{fix-cm}
\documentclass{article}       

\usepackage{graphicx}
\usepackage{mathtools}
\usepackage[noend]{algorithmic}
\usepackage{algorithm}
\usepackage{array}
\usepackage{amsmath}
\usepackage{microtype}
\usepackage{pdflscape}
\usepackage{url}
\usepackage{color}

\usepackage{epstopdf}

\begin{document}

\title{From BOP to BOSS and Beyond: Time Series Classification with Dictionary Based Classifiers}

\author{    James Large$^1$,
       Anthony Bagnall$^1$ \and
        Simon Malinowski$^2$ \and
        Romain Tavenard$^3$
}
\date{ $^1$School of Computing Sciences \\
              University of East Anglia \\
              United Kingdom\\
              ajb@uea.ac.uk\\
       $^2$  University of Rennes 1,$^3$ University of Rennes 2\\
       France\\
}

\maketitle

\begin{abstract}
A family of algorithms for time series classification (TSC) involve running a sliding window across each series, discretising the window to form a word, forming a histogram of word counts over the dictionary, then constructing a classifier on the histograms.    A recent evaluation of two of this type of algorithm, Bag of Patterns (BOP) and Bag of Symbolic Fourier Approximation Symbols (BOSS) found a significant difference in accuracy between these seemingly similar algorithms. We investigate this phenomenon by deconstructing the classifiers and measuring the relative importance of the four key components between BOP and BOSS. We find that whilst ensembling is a key component for both algorithms, the effect of the other components is mixed and more complex. We conclude that BOSS represents the state of the art for dictionary based TSC. Both BOP and BOSS can be classed as bag of words approaches. These are particularly popular in Computer Vision for tasks such as image classification. Converting approaches from vision requires careful engineering. We adapt three techniques used in Computer Vision for TSC: Scale Invariant Feature Transform; Spatial Pyramids; and Histrogram Intersection.  We find that using Spatial Pyramids in conjunction with BOSS (SP) produces a significantly more accurate classifier. SP is significantly more accurate than standard benchmarks and the original BOSS algorithm. It is not significantly worse than the best shapelet based approach, and is only outperformed by HIVE-COTE, an ensemble that includes BOSS as a constituent module.

\end{abstract}
\section{Introduction}

A family of algorithms for time series classification involve constructing a dictionary of words from the set of time series then forming a bag of words over that dictionary for each of the time series. More specifically, they run a sliding window across each series, discretise the window to form a word, form a histogram of word counts over the dictionary, then constructing a classifier on the histograms.  A recent evaluation of two of this type of algorithm, Bag of Patterns (BOP) and  Bag of Symbolic Fourier Approximation Symbols  (BOSS) found a significant difference in accuracy between these seemingly similar algorithms. We investigate this phenomena by deconstructing the classifiers and measuring the relative importance of the four key differences between BOP and BOSS. We find that ensembling makes both approaches significantly more accurate, but the effect of the other three components is more complex.

Both BOP and BOSS can be classed as bag of words approaches. These are particularly popular in Computer Vision for tasks such as image classification. Converting approaches for 2-D  image classification to 1-D series classification from a range of domains requires careful engineering. We adapt three techniques used in Computer Vision for TSC: Scale Invariant Feature Transform; Spatial Pyramids; and Histrogram Intersection.  We find that using Spatial Pyramids in conjunction with BOSS (SP) produces a significantly more accurate classifier. SP is significantly more accurate than standard benchmarks and the original BOSS algorithm. It is not significantly worse than the best shapelet based approach, and is only outperformed by HIVE-COTE, an ensemble that includes BOSS as a constituent module.

The rest of this document is structured as follows. Section~\ref{sec:background} provides an overview of the broad range of TSC algorithms, whereas Section~\ref{sec:dictionary} gives more detail into dictionary based approaches. We provide an overview of the Computer Vision framework for bag of words classification in Section~\ref{sec:refinements}. Section~\ref{sec:boptoboss} presents the results of our deconstruction of BOP and BOSS and Section~\ref{sec:bossplusplus} describes our evaluation of enhancements to BOSS. We conclude with Section~\ref{sec:conc}.

\section{TSC Background}
\label{sec:background}

A recent experimental study~\cite{bagnall17bakeoff} compared and evaluated a diverse set of twenty TSC algorithms that have been published in leading journals and conferences in the last five years. They proposed the following taxonomy of algorithms.

\subsection{Algorithms based on raw series}

Techniques based on raw series compare two series either as a vector (as with traditional classification) or by a distance measure that uses all data points. In the latter case, measures are typically combined with one-nearest-neighbour (1-NN) classifiers and the simplest variant is to compare series using Euclidean Distance. However, this baseline is easily beaten in practice, and most research effort has been directed toward finding techniques that can compensate for small misalignments between series using specialised elastic distance measures. The almost universal benchmark for whole series measures is Dynamic Time Warping (DTW) but numerous alternatives have been proposed. The most accurate whole series approach (according to the bakeoff comparison~\cite{bagnall17bakeoff}) is the Elastic Ensemble (EE)~\cite{lines15elastic}, an ensemble of 1-NN classifiers using various elastic measures, including DTW, combined through a proportional voting scheme.

\subsection{Interval-based algorithms}

Rather than use the raw series, the interval class of algorithm select one or more phase-dependent intervals of the series. At its simplest, this involves a feature selection of a contiguous subset of attributes. However, the three most effective techniques generate multiple intervals, each of which is processed and forms the basis of a member of an ensemble classifier ~\cite{deng13forest,baygogan13tsbf,baydogan15lps}. There is no significant difference in accuracy between these approaches, and the simplest is the Time Series Forest (TSF)~\cite{deng13forest}.

\subsection{Shapelet-based algorithms}

Shapelet approaches are a family of algorithms that focus on finding short patterns that define a class and can appear anywhere in the series. A class is distinguished by the presence or absence of one or more shapelets somewhere in the whole series. Shapelets were first introduced in~\cite{ye11shapelets}. The two leading ways of finding shapelets are through enumerating the candidate shapelets in the training set~\cite{lines12shapelet,hills14shapelet} or searching the space of all possible shapelets with a form of gradient descent~\cite{grabocka14invariant}. The bakeoff found that the shapelet transform algorithm used in conjunction with a heterogeneous classifier ensemble (ST-HESCA) is the most accurate approach on average.

\subsection{Dictionary-based algorithms}

Shapelet algorithms look for subseries patterns that identify a class through presence or absence. However, if a class is defined by the relative frequency of a pattern, shapelet approaches will be poor. Dictionary approaches address this by forming frequency counts of repeated patterns. They approximate and reduce the dimensionality of series by transforming into representative words, then compute similarity by comparing the distribution of words. Three of the approaches that have been published in the data mining literature are: Bag of Patterns (BOP)~\cite{lin12bagofpatterns}; the Symbolic Aggregate Approximation Vector Space Model (SAXVSM)~\cite{senin13sax_vsm}; and the Bag of Symbolic Fourier Approximation Symbols (BOSS)~\cite{schafer15boss}. We provide an overview of these algorithms in Section~\ref{sec:dictionary}.

\subsection{Spectral-based algorithms}

The frequency domain will often contain discriminatory information that is hard to detect in the time domain. Methods include constructing an autoregressive model (\cite{corduas08autoregressive,bagnall14histogram}) or combinations of autocorrelation, partial autocorrelation and autoregressive features (\cite{bagnall15cote}). An interval based spectral ensemble called Random Interval Spectral Ensemble (RISE) was proposed in~\cite{lines16hive} and shown to be more accurate on average than whole series spectral approaches.

\subsection{Combinations}
Two or more of the above approaches can  be combined into a single classifier. For example, an approach that concatenates different feature spaces is described in~\cite{kate16features}, forward selection of features for a linear classifier is the method adopted in~\cite{fulcher14comparative}) and transformation into a feature space that represents each group and ensembling classifiers together formed the basis of the Flat-COTE classifier~\cite{bagnall15cote}. A modular meta-ensemble of classifiers from each class of algorithms (EE, TSF, BOSS, ST-HESCA and RISE) called HIVE-COTE is currently the state of the art classifier for TSC when evaluated on the UCR/UEA data and simulated problems~\cite{lines16hive}. However, on individual problems, there is a wide variation between the classifiers, and the ensemble is not always the best approach. The nature of the discriminatory features will dictate the best class of algorithm.

\begin{figure}[!ht]
 	\centering
 \begin{tabular}{cc}
        \includegraphics[height =6cm, trim={2cm 3cm 2cm 2.5cm},clip]{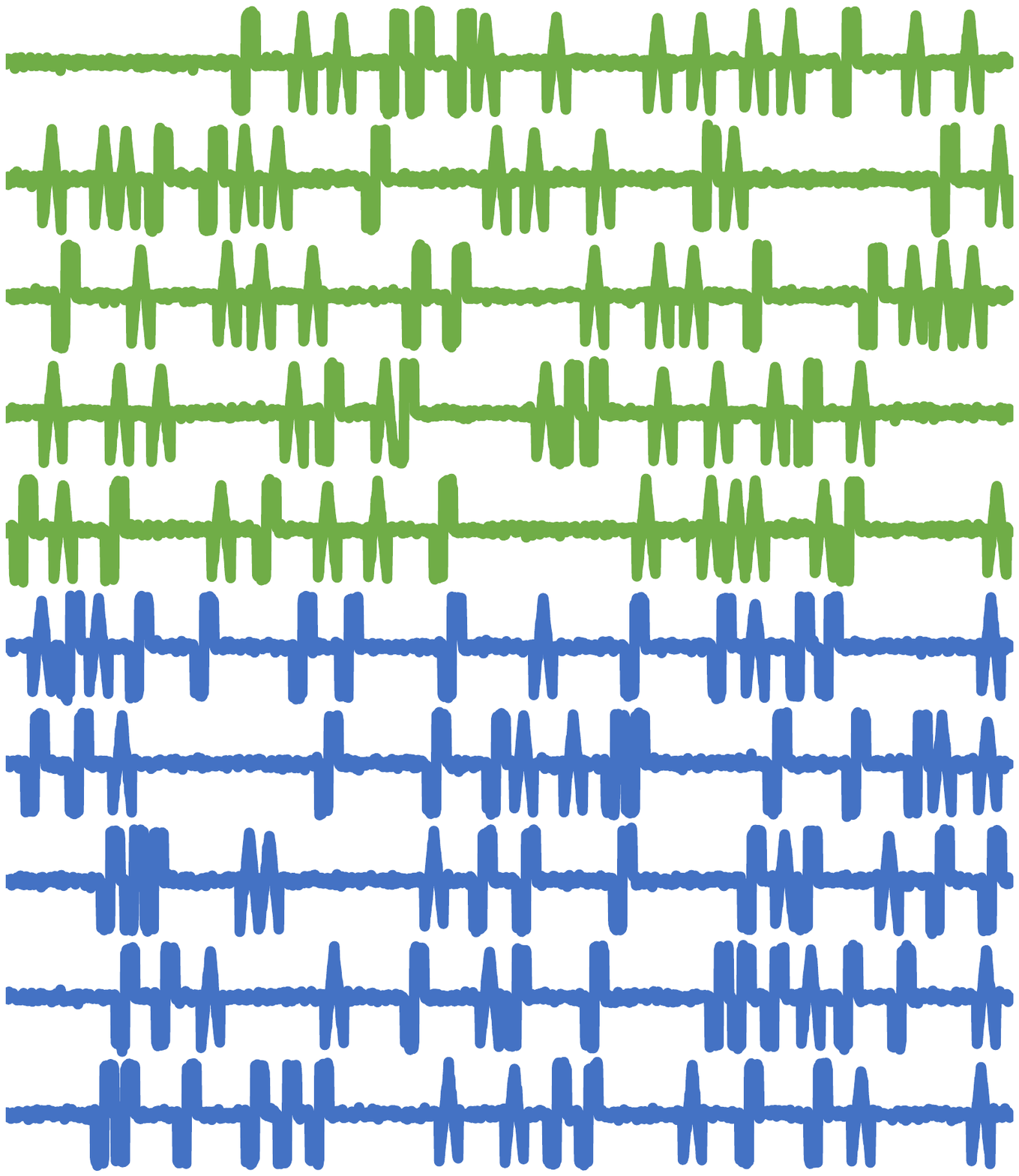}              	
 &
        \includegraphics[height=6cm,trim={2cm 3cm 2cm 2.5cm},clip]{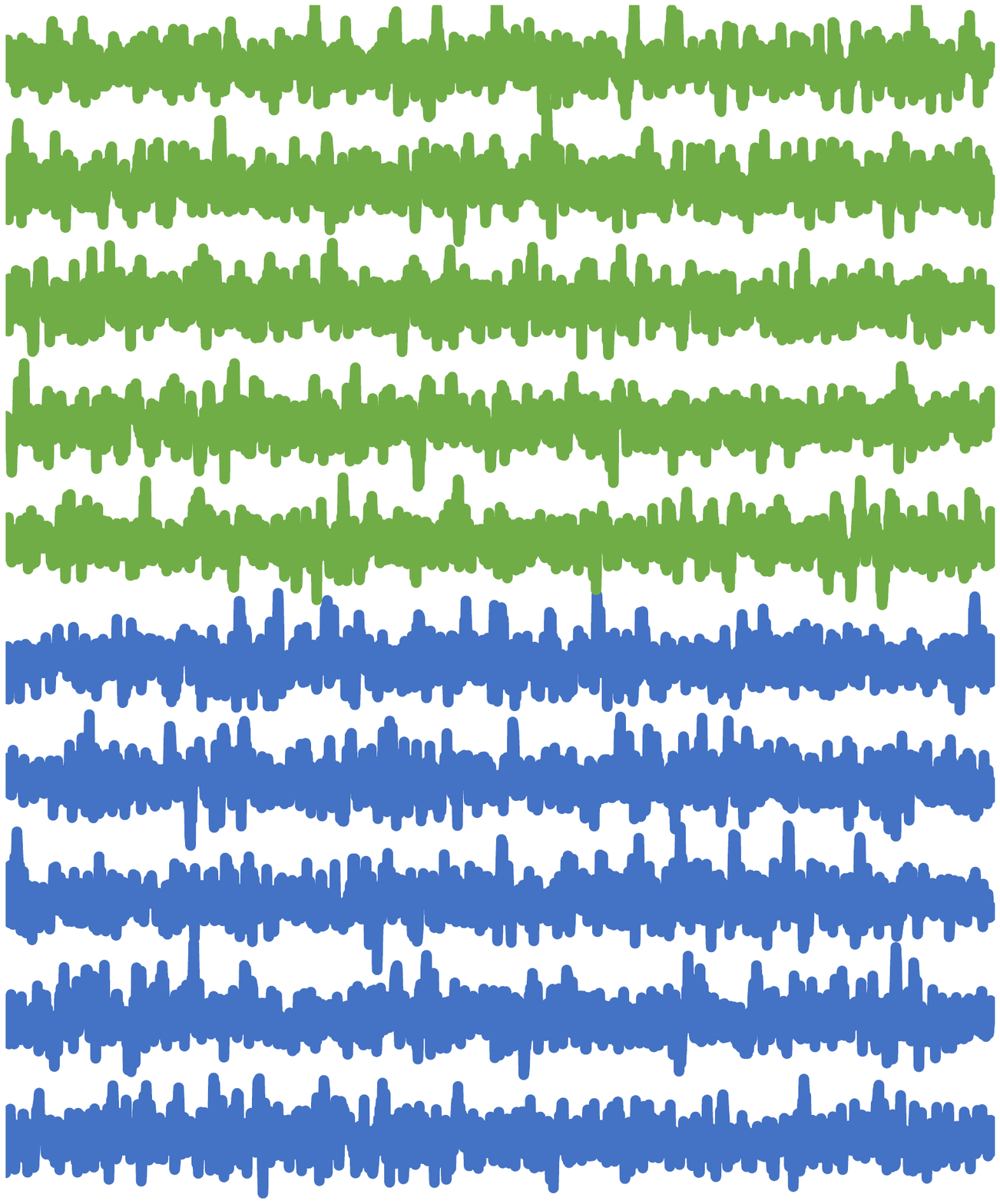}              	\\
        (a) & (b) \\
 \end{tabular}
        \caption{Two examples of simulated dictionary data.  with a mixture of truncated head and shoulders and sine shapelets.  Figure (a) has low noise and Figure (b) has standard white noise. }
        \label{dictionaryExample}
 \end{figure}

Our basic assumption is that dictionary classifiers will be best for problems where classes are defined by the frequency of occurrence of a shape in each series rather than its binary presence or absence. For example, suppose data contains short sine waves that repeat at random intervals. In one class there are many repeating patterns, in another class there are few. Figure~\ref{dictionaryExample} gives example of this kind of data.

A whole series and an interval approach will fail because the positioning of the repeating patterns is random. Shapelets will not detect this phenomena because they look for the presence or absence of a pattern. Spectral approaches may do better, but not if there are large intervals between the signals. A dictionary approach should be able to detect that one pattern occurs more frequently in one class than the other. Our objective here is to develop the best dictionary based TSC algorithm.

We describe the state of the art by summarising previously published, freely available and reproducible results\footnote{see \url{www.timeseriesclassification.com} for details}. We compare the relative performance of three base line classifiers: rotation forest with 50 trees (RotF); 1-NN with Euclidean distance (Euclid); DTW with window set through cross validation (DTW), a representative of each class of algorithm: EE, TSF, ST and RISE, the three dictionary classifiers BOP, SAXVSM and BOSS and two ensemble approaches, Flat-COTE and HIVE-COTE.

\begin{figure}[!ht]
	\centering
    \includegraphics[width=\linewidth,trim={3cm 10cm 3cm 12.5cm},clip]{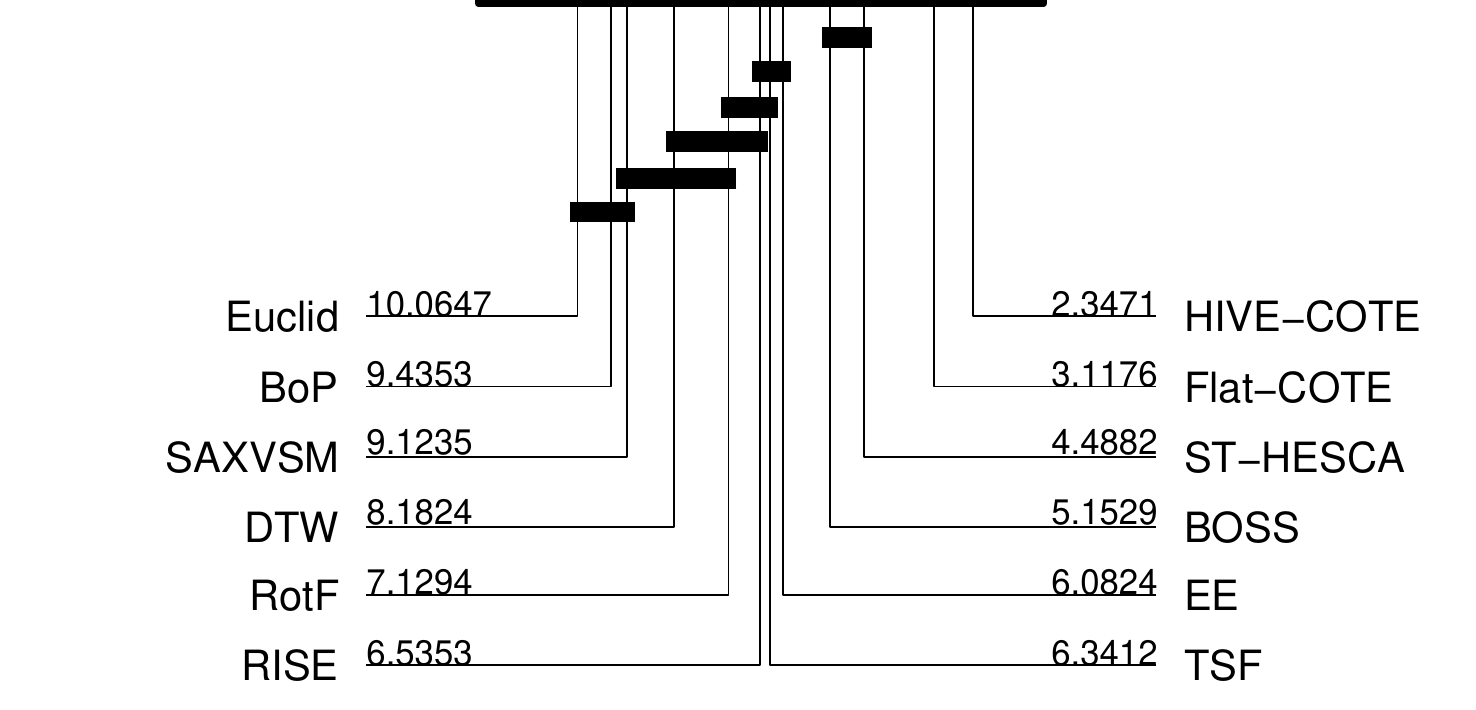}
       \caption{Average ranks of 12 classifiers on 100 resamples of 85 data sets. The results were first presented in~\cite{bagnall17bakeoff} and~\cite{lines18hive}. A solid bar across a set of classifiers indicates there is no significant difference within that group.}
       \label{bakeoffResults}
\end{figure}

To compare multiple classifiers on multiple problems,  following the recommendation of Dem\v{s}ar \cite{demsar06comparisons}, we use the Friedmann test to determine if there were any statistically significant differences in the rankings of the classifiers. However, following recent recommendations in \cite{benavoli16pairwise} and \cite{garcia08pairwise}, we have abandoned the Nemenyi post-hoc test originally used by
\cite{demsar06comparisons} to form cliques (groups of classifiers within which there is no
significant difference in ranks). Instead, we compare all classifiers with pairwise Wilcoxon
signed rank tests, and form cliques using the Holm correction (which adjusts family-wise
error less conservatively than a Bonferroni adjustment).

HIVE-COTE is the most accurate algorithm over all, but features of these results excited our interest about dictionary classifiers. Firstly, BOP and SAXVSM performed very poorly. Neither is significantly better than 1-NN Euclidean distance and neither could beat the benchmark classifiers rotation forest and DTW. In stark contrast, BOSS is one of the best performers. It is not significantly worse than the ST-HESCA and only beaten by the two meta ensembles Flat-COTE and HIVE-COTE. HIVE-COTE contains BOSS whereas Flat-COTE does not, and the fact that HIVE-COTE is significantly better than Flat-COTE is further evidence in support of BOSS. On a head to head comparison, BOSS beats BOP on 80 of the 85 datasets. The mean difference in accuracy is over 8\%. These algorithms are seemingly similar, so why is BOSS so much better than BOP? Answering this question requires a more in depth understanding of how these algorithms work.

\section{Dictionary Based Algorithms}
\label{sec:dictionary}


Dictionary based algorithms share the same basic structure. In summary, a window of length $w$ is passed across each series. Each subseries is then represented by some string or pattern that is representative of it. In the cases considered here, each subseries from the windowing is first compressed from length $w$ to $l$. The shortened subseries are then discretised, so that each of the $l$ data is restricted to one of $\alpha$ values. The occurrence of the resulting `word', ${\bf r}$, is recorded in a histogram, although in a stage called numerosity reduction, contiguous series of identical words are counted as a single occurrence. Each series has a separate histogram (also referred to as bag), and new instances are classified based on the distance between their own histogram and those in the training set, based by default on 1-nearest neighbour classification, though other methods could be used.


There are four key stages at which major differences between dictionary based algorithms may arise:
\begin{enumerate}
\item the compression method to get from $w$ real valued data to $l$ real valued data;
\item the discretisation technique used to convert the $l$ real valued data into $l$ discrete data with $\alpha$ possible values;
\item the methods of representing the collections of transformed subseries; and
\item the distance measure used to compare histograms and/or the classification algorithm used to classify new cases.
\end{enumerate}

\subsection{Bag of Patterns (BOP)~\cite{lin12bagofpatterns}}

BOP (described in Algorithm~\ref{bop}) is a dictionary classifier built on the Symbolic Aggregate Approximation (SAX)~\cite{lin07sax} algorithm. SAX reduces $w$ to $l$ through Piecewise Aggregate Approximation (PAA) (i.e. each of the $l$ new points is an average over an interval length $w/l$) and discretises to $\alpha$ values using quantiles of the normal distribution. If consecutive windows produce identical words,  then only the first of that run is recorded. This is included to avoid the over counting of trivial matches, especially in smooth regions of the originating series. The distribution of words over a series forms a count histogram. To classify new samples, the same transform is applied to the new series and the nearest neighbour histogram within the training matrix found. BOP sets the three parameters through cross validation on the training data.

\begin{algorithm}[!ht]
	\caption{buildClassifierBOP(A list of $n$ cases of length $m$, ${\bf T}=({\bf X,y})$)}
	\label{bop}
	\begin{algorithmic}[1]
\REQUIRE the word length $l$, the alphabet size $\alpha$ and the window length $w$
		\STATE Let ${\bf H}$ be a list of $n$ histograms $({\bf h}_1,\ldots,{\bf h}_n)$
		\STATE ${\bf p} \leftarrow \emptyset$
		\FOR {$i \leftarrow  1$ to $n$}
			\FOR {$j \leftarrow 1$ to $m-w+1$}
				\STATE ${\bf q} \leftarrow x_{i,j} \ldots x_{i,j+w-1}$
				\STATE ${\bf r} \leftarrow $ SAX($q, l, \alpha$)
				\IF{${\bf r} \neq {\bf p}$}
					\STATE $pos \leftarrow$ index(${\bf r}$) \COMMENT{{\em the function} index {\em determines the location of the word ${\bf r}$ in the count matrix ${\bf h_i}$}}
					\STATE ${h}_{i,pos} \leftarrow {h}_{i,pos} + 1$
				\ENDIF
				\STATE ${\bf p} \leftarrow {\bf r}$
			\ENDFOR
		\ENDFOR
	\end{algorithmic}
\end{algorithm}

The Symbolic Aggregate Approximation - Vector Space Model (SAXVSM)~\cite{senin13sax_vsm} combines the SAX representation used in BOP with the Vector Space Model commonly used in Information Retrieval. The key differences between BOP and SAXVSM is that SAXVSM forms word distributions over classes rather than series and weights these by the term frequency/inverse document frequency ($tf\cdot idf$). For SAXVSM, term frequency $tf$ refers to the number of times a word appears in a class and document frequency $df$ means the number of classes a word appears in. $tf\cdot idf$ is then defined as
$$\begin{displaystyle}
tfidf(tf, df) = \left.
\begin{cases}
\log{(1+tf)}\cdot \log(\frac{c}{ df }) & \text{if } df > 0 \\
0 & otherwise \\
\end{cases} \right.
\end{displaystyle}$$
where $c$ is the number of classes. There is no significant difference in accuracy between BOP and SAXVSM, so we can without loss of generality restrict our attention to BOP.

\subsection{Bag of Symbolic Fourier Approximation Symbols (BOSS)~\cite{schafer15boss}}

BOSS also uses windows to form words over series, and represents them in a simple histogram format, but it has several major differences to BOP and SAXVSM. BOSS uses a truncated Discrete Fourier Transform (DFT) instead of a PAA on each window. Another difference is that the truncated series is discretised through a technique called Multiple Coefficient Binning (MCB), rather than using fixed intervals. MCB finds the discretising break points as a preprocessing step by estimating the distribution of the Fourier coefficients. This is performed by segmenting the series into disjoint windows, performing a DFT, then finding breakpoints for each coefficient such that each bin contains the same number of elements. The whole process of forming words is called Symbolic Fourier Approximation (SFA). BOSS then involves similar stages to BOP; it windows each series to form the term frequency through the application of DFT and discretisation by MCB, performs numerosity reduction, and forms histograms of the words in each series. A bespoke distance function is used for nearest neighbour classification. This non symmetrical function only includes distances between frequencies of words that actually occur within the first histogram passed as an argument, which refers to the test case. 

Another major difference is that BOSS forms an ensemble by retaining all classifiers with training accuracy within 92\% of the best during the parameter search of window sizes. New instances are classified by a majority vote of the resulting ensemble. Algorithm \ref{boss} details the construction of histograms for a given parameter set.

\begin{algorithm}[!ht]
	\caption{buildClassifierBOSS(A list of $n$ cases of length $m$, ${\bf T}=({\bf X,y})$)}
	\label{boss}
	\begin{algorithmic}[1]
\REQUIRE the word length $l$, the alphabet size $\alpha$, the window length $w$, normalisation parameter $p$
		\STATE Let ${\bf H}$ be a list of $n$ histograms $({\bf h}_1,\ldots,{\bf h}_n)$
		\STATE Let ${\bf B}$ be a matrix of $l$ by $\alpha$ breakpoints found by MCB
		\STATE ${\bf p} \leftarrow \emptyset$
		\FOR {$i \leftarrow  1$ to $n$}
			\FOR {$j \leftarrow 1$ to $m-w+1$}
				\STATE ${\bf s}\leftarrow x_{i,j} \ldots x_{i,j+w-1}$
				\STATE ${\bf q} \leftarrow$ DFT(${\bf s}, l, \alpha$,$p$) \COMMENT{ {\em {\bf q} is a vector of the complex DFT coefficients}}
				\STATE ${\bf q'} \leftarrow (q_1 \ldots q_{l/2})$
				\STATE ${\bf r} \leftarrow$ MCB(${\bf q', B}$)
				\IF{${\bf r} \neq {\bf p}$}
					\STATE $pos \leftarrow $index(${\bf r}$)
					\STATE ${h}_{i,pos} \leftarrow {h}_{i,pos} + 1$
				\ENDIF
				\STATE ${\bf p} \leftarrow {\bf r} $
			\ENDFOR
		\ENDFOR
	\end{algorithmic}
\end{algorithm}

In a manner reminiscent of the way SAXVSM adapts BOP, BOSS-Vector Space (BOSS-VS)~\cite{schafer16boss-vs} modifies BOSS to form class histograms rather than instance histograms. Switching to class histograms massively reduces the memory requirements and speeds up classification, but it has no significant effect on accuracy, unless it is to reduce it (see the results in~\cite{schafer16boss-vs}). In this work we are concerned with classification accuracy. The questions we address are, firstly, why is BOSS so much better than BOP (see Section~\ref{sec:boptoboss}) and secondly, can we refine BOSS to make it more accurate (see Section~\ref{sec:bossplusplus}).

\section{Computer Vision Bag of Words Framework}
\label{sec:refinements}
The histogram approach used by dictionary classifiers has similarities to many approaches used in the field of Computer Vision. A typical Computer Vision bag of words framework involves the following stages:
\begin{enumerate}
\item  extraction of keypoints;
\item description of keypoints;
\item bag forming; and
\item classification based on bags.
\end{enumerate}
BOP and BOSS extract keypoints through sliding a window over the whole series, reducing the size of the number of keypoints through numerosity reduction and a restriction of window sizes. However, approaches for dictionary based TSC more in line with the Computer Vision approach have been proposed. \cite{bailly16botsw} describes an approach for using Scale Invariant Feature Transform (SIFT) \cite{lowe04sift} features for use in TSC with dictionary classifiers. We describe this approach in detail in Section~\ref{botsw} and have implemented a version in the WEKA based TSC codebase\footnote{\url{https://bitbucket.org/TonyBagnall/time-series-classification}}.

We also consider a common technique in Computer Vision called Spatial Pyramids, proposed in \cite{lazebnik06beyond} and described in Section~\ref{sp}. We try incorporating this as a wrapper for BOSS. It could equally be applied to other dictionary approaches.

A more complex Computer Vision approach applied to TSC is proposed in~\cite{zhao16descriptors}. This involves a combined approach of peak finding and hybrid sampling to extract keypoints, using Histogram of Oriented Gradients (HOG-1D) and Dynamic Time Warping-Multidimensional Scaling (DTW-MDS) to form features describing the keypoints, clustering them with a $K$ component Gaussian Mixture Model, forming bags based on Fisher Vector encoding and finally constructing a linear kernel Support Vector Machine classifier. The resulting classifier, called HOG-1D+DTW-MDS, is evaluated on the standard single folds of 43 of the UEA/UCR data sets. They do not compare the results of HOG-1D+DTW-MDS to the published results for BOSS, presumably due to the lag time in publication. Using the results in Table 3 from~\cite{zhao16descriptors} and the BOSS results presented in~\cite{bagnall17bakeoff}, we find no significant difference between HOG-1D+DTW-MDS and BOSS (HOG-1D+DTW-MDS wins on 22, BOSS on 19 and they tie on 2).


\subsection{Bag of Temporal SIFT Words (BOTSW) Classifier}
\label{botsw}

The Bag of Temporal SIFT Words (BOTSW) algorithm~\cite{bailly16botsw} adopts a version of the Computer Vision bag of words framework that is easier to reproduce than that described in~\cite{zhao16descriptors}, not least because the C++ source code is publicly available\footnote{\url{https://github.com/a-bailly/dbotsw}}.
BOTSW first extracts keypoints from every time series through regular sampling at a rate $r$, which is a parameter of the method. Then, each keypoint is described by $n_s$ feature vectors, where $n_s$ is the number of considered scales. Each feature vector describes the keypoint at a particular scale. To obtain the feature vector of a keypoint at a scale $s$, the time series is filtered by a Gaussian filter of width $s$. $n_b$ blocks of size $\alpha$ are selected around the keypoint. Gradients of the filtered time series are computed for every point of every block and then weighted by a Gaussian function to give greater importance to those points nearer to the keypoint.

Each block is described by two values: the sum of positive gradients in the block and the sum of negative gradients. A feature vector that describes a keypoint at a particular scale is a 2$n_b$-long vector.
A dictionary of feature vectors is learned by a $k$-means clustering on the whole set of feature vectors from the time series database. Feature vectors are then quantized using the dictionary.
The number of occurrences of these words in the series is computed to form a histogram, which is normalised using Signed Square Root (SSR) then $l_2$ normalisation. This nomalized histogram is the final feature vector representing this series.

In \cite{bailly16botsw}, a Support Vector Machine was used to classify feature vectors. However, our objective is to assess the utility of the SIFT features in relation to the BOSS features. Hence, to minimize the differences between BOSS and BOP we use 1NN classification. The parameters $n_b$ in \{4, 8, 12, 16, 20\}, $\alpha$ in \{4, 8\}, and $k$ in \{32, 64, 128, 256, 512, 1024\} are tuned through a grid search with cross validation. To further align with BOSS, we form an ensemble of BOTSW classifiers, retaining all parameter sets with training accuracy within 92\% of the global maximum. This  homogeneous ensemble classifies new instances with a simple majority vote.




\subsection{BOSS Ensemble with Spatial Pyramids (SP)}
\label{sp}

The essence of dictionary classifiers is to ignore temporal information through consideration of the recurrence of short subseries. Whilst this will lead to good results in problem domains with repeated discriminatory features, the disadvantage is that in some domains the location in time of a pattern is as important as the pattern itself. Spatial pyramids \cite{lazebnik06beyond} are a method commonly used in Computer Vision, which will allow us combine temporal and phase independent features. When applied to time series, using a spatial pyramid involves recursively segmenting each series and constructing histograms on the segments.

Starting from the initial histogram across the whole series, histograms on subsections are formed by repeatedly dividing the series $L$ times. These histograms are weighted by $\frac{1}{2^{L-l}}$, which is inversely proportional to the level $l$ at which they are found. All histograms are then combined and normalised to form a single elongated histogram feature.

Because of the weighting, similarity between features found at smaller divisions on the series have a more significant effect than those found on a more global scale, as their temporal location becomes increasingly dissimilar. It is also worth noting that a pyramid with one level is equivalent to the basic bag of words, as no division has occurred.

\begin{figure}[!ht]
	\centering
    \includegraphics[width=\linewidth]{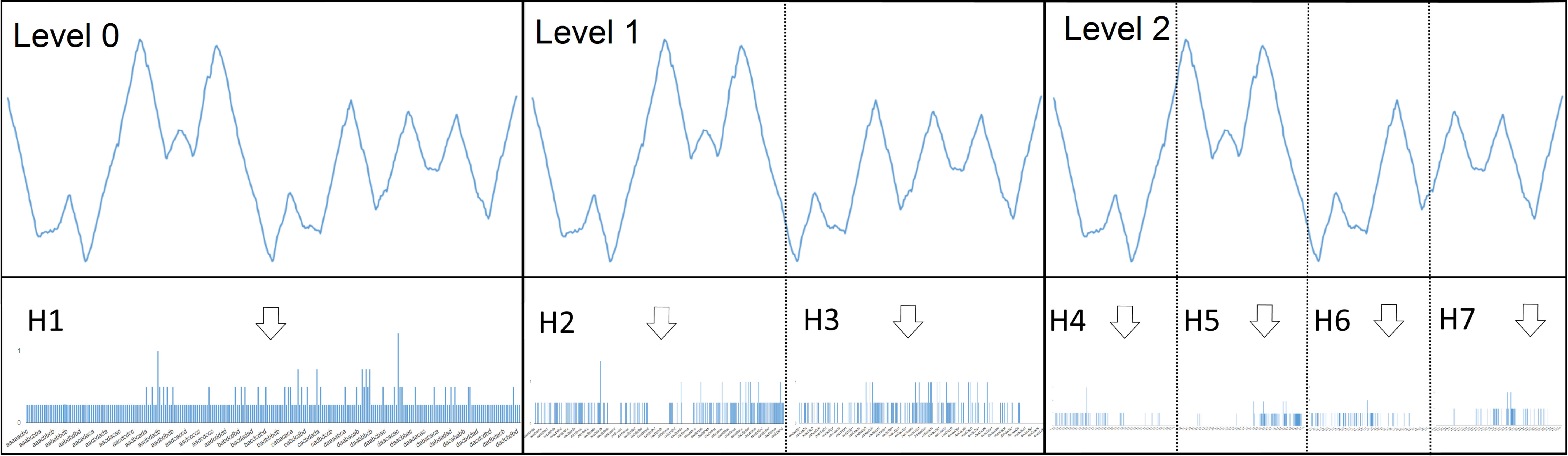}
       \caption{A series from the BeetleFly dataset, being divided at successive levels with Bags of SFA words being formed for each subsection. H1...7 are combined to form the final feature vector.}
       \label{fig:bosssp}
\end{figure}

Since BOSS ensembles over different window sizes so that discriminatory patterns of different lengths can all be accounted for, we search for $L$ for each member of the ensemble during training. An overview of the ensemble construction is given in Algorithm \ref{alg:bossSP}. Feature sets formed from an optimal word length, found through CV, for a given window size are generated as usual. This feature set, implicitly produced as a pyramid with $L=1$, is then augmented and further CV is performed to find the optimal $L$ in \{1,2,3\}. This effectively defines whether the discriminatory feature type described by this parameter set is more local or global in nature. If the training accuracy of the best word length and number of levels for this window size falls within 92\% of the best, it is included in the ensemble. In classification, for each member the test instance is transformed into a spatial pyramid using that member`s parameters, and the class of the train pyramid with the maximal Histogram Intersection or minimal BOSS Distance is returned. Figure~\ref{fig:bosssp} gives an example of the process of forming histograms for SP.

        \begin{algorithm}[!ht]
        \caption{buildBOSSEnsembleSP(A list of $n$ cases of length $m$, ${\bf T}=\{{\bf X,y}\}$)}
        \label{alg:bossSP}
        \begin{algorithmic}[1]
            \STATE $\alpha$ = 4
            \STATE featureSets = [features, trainAccuracies]
            \FOR {w in windowLengths()}
                \STATE bestWindowFeatureSet = null
                \STATE bestWindowAcc = 0

                \FOR {k in wordLengths()}
                    \STATE featureSet = BOSSTransform(w,k,$\alpha$)
                    \STATE acc = CrossValidate(featureSet)
                    \IF {acc $>$ bestWindowAcc}
                        \STATE bestWindowAcc = acc, bestWindowFeatureSet = featureSet
                    \ENDIF
                \ENDFOR

                \FOR {L in {2,3}}
                    \STATE featureSet = buildPyramid(bestWindowFeatureSet, L)
                    \STATE acc = CrossValidate(featureSet)
                    \IF {acc $>$ bestWindowAcc}
                        \STATE bestWindowAcc = acc, bestWindowFeatureSet = featureSet
                    \ENDIF
                \ENDFOR

                \STATE featureSets.add(bestWindowFeatureSet, bestWindowAcc)
            \ENDFOR
            \STATE maxWindowAcc = max(featureSets.trainAccuracies)
            \FOR {set in featureSets}
                \IF {bestWindowAcc $>$ maxWindowAcc*0.92}
                    \STATE addToEnsemble(set)
                \ENDIF
            \ENDFOR
        \end{algorithmic}
    \end{algorithm}

    While computing the pyramids is very fast relative to the original production of the SFA words, the additional space complexity is a concern for large datasets as the final elongated histograms will be $\sum_{l=0}^{L-1} 2^l$ times larger. This can be heavily mitigated by using sparse data representations, since histograms at higher levels will be more sparse than those at lower levels. However, a cap of 100 was also placed on the maximum size of the ensemble to keep the space requirements more reasonable. Thus if $\lambda$ is the number of feature sets within the threshold of the max accuracy, the size of the final ensemble is $min(100, \lambda)$.

\subsection{Histogram Intersection (HI) Distance}

A core task in any bag of words/dictionary based technique is to compare the differences between the resulting histograms in order to define class membership. BOP uses Euclidean Distance, SAX-VSM uses Cosine Similarity, and BOSS its own measure which is a slight alteration to Euclidean. We also test the Histogram Intersection similarity measure described in \cite{lazebnik06beyond} which is used in many different applications involving histograms. For a dictionary and resulting histogram size of $k$, this is defined as:

        \begin{center}
            $\begin{displaystyle}
            HI(\textbf{a}, \textbf{b}) = \sum_{i=1}^{k} min(a_i, b_i)
            \end{displaystyle}$
        \end{center}

\section{From BOP to BOSS}
\label{sec:boptoboss}

We perform all experiments using 77 of the datasets at the University of California, Riverside/University of East Anglia (UCR/UEA) time series classification repository\footnote{UCR/UEA TSC Repository: \url{ www.timeseriesclassification.com}} (\cite{bagnall17bakeoff}). There are 8 datasets that we do not use for practical reasons: their size means the classifiers take too long to train or require too much memory to complete given our time frame and the number of experiments and resampling performed. The full list of problems we used is given in Table~\ref{tab:results}. Our focus is on bridging the accuracy gap between the two classifiers; optimizing for speed and memory are of course very important, but are not the focus of this study. All of our code and data is available from a public code repository and accompanying website\footnote{\url{www.timeseriesclassification.com/dami2017.php}}. We compare classifiers by the accuracy average over 25 stratified resamples (with the same train/test size of the original data).

Both BOP and BOSS tune their parameters through a leave one out cross validation on the train data for a predefined parameter space. The results for BOSS and BOP presented in~\cite{bagnall17bakeoff} were obtained using the parameter space defined in the original papers, and these parameter spaces are different. To alleviate this possible source of bias we have altered the BOP search space to match that of BOSS (see Table~\ref{tab:altparamspaces}). In a pairwise comparison between the $BOP$ on the old and new parameter space, the latter had higher average accuracy on 44 datasets and worse on 33. There is no significant difference between the old and new versions, and we conclude that we cannot explain the difference between BOP and BOSS on this factor.
				
		\begin{table}[!htbp]
			\centering
			\caption{Parameter search spaces for BOP and BOSS.}
			\begin{tabular}{c|c}
				\hline
				Algorithm & Parameters \\
				\hline
				$BOP$ published parameter search space    & $w$ = {15\%\dots 36\% of m}\\
                & $l$ = powers of 2 up to w/2\\
                & $\alpha$ = {2\dots8}    \\ \hline
				$BOSS$ published parameter search space  & $w$ = 10\dots m\\
This space is used for both BOSS and BOP                &$l$ = {8, 10, 12, 14, 16}\\
                &$\alpha$ = 4    \\
				\hline
			\end{tabular}
			\label{tab:altparamspaces}
		\end{table}
BOP and BOSS are identical except for four features.
\begin{enumerate}
\item {\bf The window approximation method}. Each window of length $w$ is reduced to a series of length $l$ through an approximation method. BOP uses Piecewise Aggregate Approximation whereas BOSS uses the truncated Fourier terms.
\item {\bf The discretisation method}. Each value in the approximate series of length $l$ is discretised into one of $\alpha$ values. BOP uses the fixed intervals defined in SAX whereas BOSS uses the data driven technique Multiple Coefficient Binning (MCB).
\item {\bf The distance measure}. BOP uses 1-NN with Euclidean distance whereas BOSS uses a 1-NN with a bespoke, non-symmetric distance function that ignores zero entries in the test histogram.
\item {\bf The classifier}. BOSS is an ensemble of multiple transforms with different parameters, whereas BOP is a single classifier.
\end{enumerate}
	
To quantify the source of the difference in BOP and BOSS, we assess the relative importance of each of these components by adding each of the four BOSS features into BOP. We then measure their importance to BOSS by in turn replacing each BOSS feature with that used in BOP. This gives us the 10 BOP/BOSS variants listed in Table~\ref{tab:alterations}.
				
		\begin{table}[!htbp]
			\centering
			\caption{BOP and BOSS variants, with the switching of BOSS and BOP features. For  clarity, we indicate the variant by the addition or removal of the boss feature. e.g. BOP+Ens is BOP with an added BOSS-like ensemble, whereas BOSS-BD is BOSS with Euclidean distance replacing BOSS distance.}
			\begin{tabular}{l|c}
				\hline
				Algorithm & Label \\
				\hline
				Base BOP classifier & $BOP$\\
				BOP with FT approximation replacing PAA & $BOP+FT$\\
				BOP with MCB discretisation replacing Gaussian breakpoints& $BOP+MCB$\\
				BOP with BOSS Distance measure replacing Euclidean Distance  & $BOP+BD$\\
				BOP with ensembling over best parameter sets & $BOP+Ens$\\
				\hline
				Base BOSS classifier & $BOSS$\\
				BOSS with PAA approximation replacing FT& $BOSS-{FT}$\\
				BOSS with Gaussian breakpoint discretisation replacing FT& $BOSS-{MCB}$\\
				BOSS with Euclidean Distance metric replacing BOSS Distance& $BOSS-{BD}$\\
				BOSS with single best parameter set & $BOSS-{Ens}$\\
				\hline
			\end{tabular}
			\label{tab:alterations}
			
		\end{table}
	
\begin{figure}[!ht]
	\centering
    \includegraphics[width=\linewidth,trim={3cm 11cm 3cm 10cm},clip]{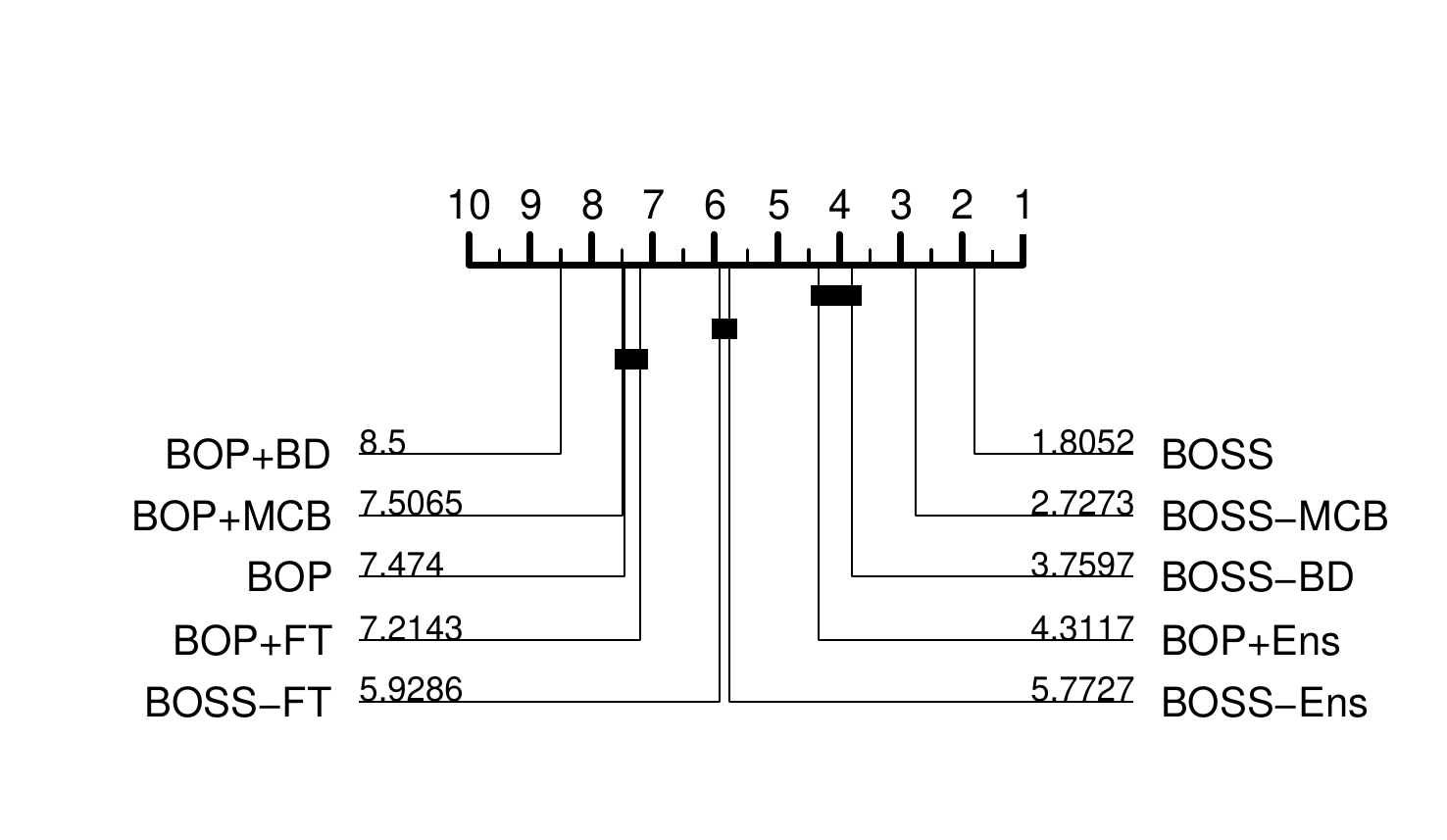}
       \caption{Average ranks and cliques of 10 BOP/BOSS classifiers on 25 resamples of 77 data sets.}
       \label{fig:boptoboss}
\end{figure}

Figure~\ref{fig:boptoboss} shows the average ranks of the 10 variants of BOP and BOSS we have evaluated. Table \ref{tab:meanDiffs} shows the mean difference in accuracy between the variants.  The mean difference between the best and the worst variant is nearly 9\%. To be clear, this is the absolute pairwise difference in accuracy; the worst algorithm, BOP+BD has an average accuracy over all problems of 73.8\%, whereas BOSS, the best algorithm, has average accuracy of 82.65\%.

		\begin{table}[!htbp]
			\centering
			\caption{The mean difference in accuracy between BOP and BOSS variants over 77 datasets.}
          \scalebox{0.7}{
			\begin{tabular}{ccccccccccc}
	& BOP+BD	& BOP+MCB &	BOP	& BOP+FT & BOSS-FT & BOSS-Ens &BOP+Ens &	BOSS-BD	& BOSS-MCB &	BOSS \\ \hline
BOP+BD      & 0.00\% & 	-0.97\%  &-0.97\%   & -0.71\%&	-3.10\%	& -3.61\%	& -4.94\%   & 	-5.59\% &-7.60\% &	-8.88\%  \\
BOP+MCB     & 0.97\% & 	0.00\%	 &-0.01\%   & 0.26\% & 	-2.13\%	& -2.65\%	& -3.97\%	& -4.62\%	&-6.63\% &	-7.91\%  \\
BOP         & 0.97\% & 	0.01\%	 & 0.00\%   & 0.26\% & 	-2.12\%	& -2.64\%	& -3.96\%	& -4.61\%	&-6.63\% &	-7.91\%  \\
BOP+FT      & 0.71\% & 	-0.26\%  &-0.26\%   & 0.00\% &	-2.39\%	& -2.90\%	& -4.23\%	& -4.88\%	&-6.89\% &	-8.17\%  \\
BOSS-FT     & 3.10\% & 	2.13\%	 & 2.12\%   & 2.39\% & 	0.00\%  & -0.51\%	& -1.84\%	& -2.49\%	&-4.50\% &	-5.78\%  \\
BOSS-Ens    & 3.61\% & 	2.65\%	 & 2.64\%   & 2.90\% & 	0.51\%  & 	0.00\%	& -1.32\%	& -1.98\%	&-3.99\% &	-5.27\%  \\
BOP+Ens     & 4.94\% & 	3.97\%	 & 3.96\%   & 4.23\% & 	1.84\%  & 	1.32\%	& 0.00\%	& -0.65\%	&-2.66\% &	-3.94\%  \\
BOSS-BD     & 5.59\% & 	4.62\%	 & 4.61\%   & 4.88\% & 	2.49\%  & 	1.98\%	& 0.65\%	& 0.00\%	&-2.01\% &	-3.29\%  \\
BOSS-MCB    & 7.60\% & 	6.63\%	 & 6.63\%   & 6.89\% & 	4.50\%  & 	3.99\%	& 2.66\%	& 2.01\%	&0.00\%	 & -1.28\%  \\
BOSS        & 8.88\% & 	7.91\%	 & 7.91\%   & 8.17\% & 	5.78\%  & 	5.27\%	& 3.94\%	& 3.29\%	&1.28\%	 & 0.00\%   \\
			\end{tabular}
}
			\label{tab:meanDiffs}			
		\end{table}

We can make the following observations from these results. For BOP, using DFT with fixed boundaries, or PAA with MCB discretisation, makes no difference to using SAX. This suggests the benefit of using spectral features only comes when using bespoke bins to discretise. Using BOSS distance for BOP histograms actually makes BOP significantly worse. The only component of BOSS that significantly improves BOP is ensembling. This actually makes BOP significantly better than a single version of BOSS (the mean difference 1.32\%). However, BOP ensemble is still significantly worse than the BOSS ensemble. Hence we cannot attribute the difference to the ensembling method alone: removing any one of the four components of BOSS makes it significantly worse. The worst change to make is to switch from DFT features to PAA features. This surprised us, as we had assumed removing ensembling would cause the most harm. Clearly the four features of BOSS that differentiate it from BOP are all required and all interact to produce a better classifier. This highlights that the engineering of algorithms is often not linear, and components can interact in ways that may not be intuitively obvious. This is most clearly observable when comparing the effect of using the BOSS Distance (BD): BD makes BOP significantly worse but BOSS significantly better.


There is some difference in the structures of the resulting histograms of each transform which BOSS Distance is able to leverage over Euclidean Distance in the case of BOSS, but not BOP. Considering the actual bagging process is essentially the same between the two - both extract words from a massively sparse space, and both use numerosity reduction - such an apparently clear, or rather `usable', distinction in the final histograms is striking. BOSS Distance ignores words that do not appear in the test histogram, so for BOP these missing words are seemingly informative, however for BOSS they are noise and removing them is beneficial. Exactly why this is so is unclear. We suspect this difference is due to the action of MCB, which creates a data driven discretisation. If the underlying distribution diverges significantly from normality, MCB will create a more accurate representation and will hence capture an underlying pattern more accurately. This could lead to the truly informative words being separated from the uninformative noise more successfully, and so the words that are ignored are more likely to be noise.
This is just conjecture. Further work and analysis of the resulting histograms would need to be performed to fully understand the mechanisms at work here,
however this is beyond the scope of this work.

From these experiments we conclude that BOSS, as described in~\cite{schafer15boss}, represents the state of the art for dictionary classifiers that were first introduced in~\cite{lin12bagofpatterns}. Our next question is, can we improve on the state of the art?

\section{BOSS Extensions}
\label{sec:bossplusplus}
In Section~\ref{sec:refinements} we described two approaches from Computer Vision that may improve dictionary classifiers: SIFT features~\cite{lowe04sift}, adapted for time series as described in~\cite{bailly16botsw} and Spatial Pyramids~\cite{lazebnik06beyond} that have not formerly been used in this context. We also  described the Histogram Intersection as an alternative distance measure between histograms. We wish to assess whether adding these alternative structures to the state of the art dictionary classifier gives an overall improvement in accuracy. To try and isolate the causes of any observed differences we start with BOSS as the base classifier and make the minimum adjustments.

\begin{figure}[!ht]
	\centering
    \includegraphics[width=\linewidth,trim={3cm 13cm 3cm 11cm},clip]{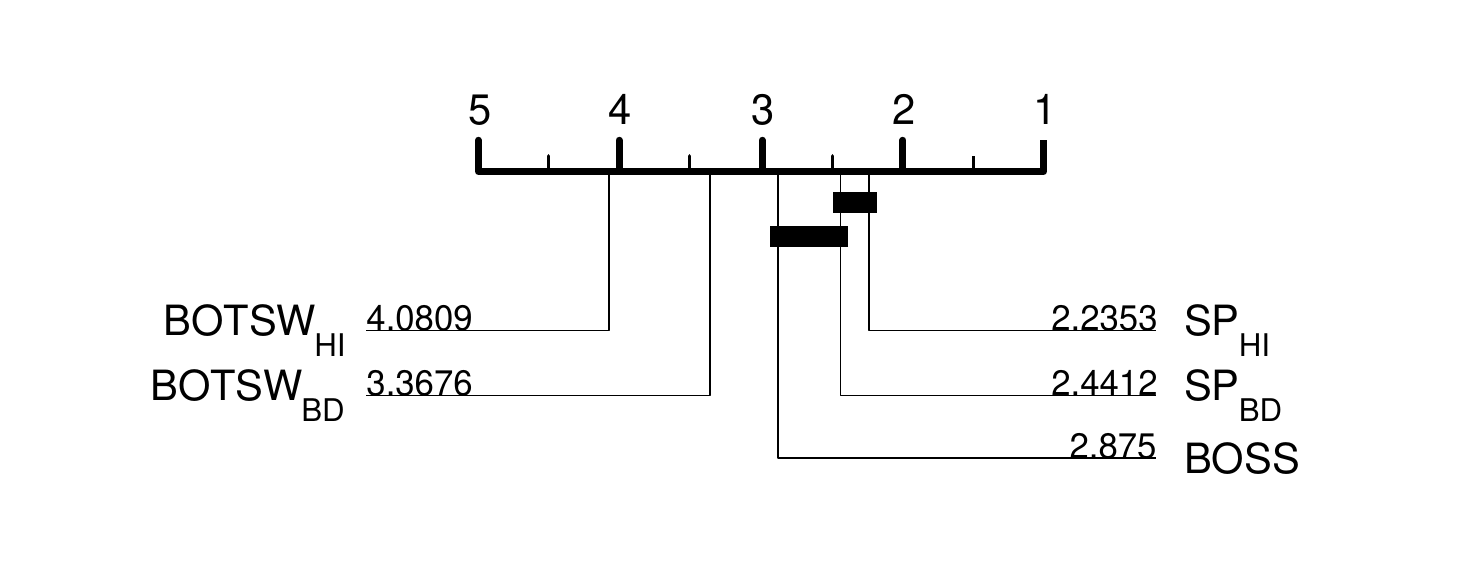}
       \caption{Average ranks and cliques for five variants of dictionary classifiers.}
       \label{fig:beyondboss}
\end{figure}

SP$_{BD}$ is a spatial pyramid built on top of the standard BOSS ensemble (using BD).  SP$_{HI}$ is a spatial pyramid built on top of BOSS, using histogram intersection. BOTSW$_{BD}$ is a bag of temporal-sift classifiers that use BD, whereas BOTSW$_{HI}$ uses histogram intersection. All four classifiers ensemble in an identical way to BOSS (retain all models within 92\% of the best), and each pair of classifiers (i.e. BOTSW$_{BD}$/BOTSW$_{HI}$ and SP$_{BD}$/SP$_{HI}$) search identical parameter spaces. The average ranks and cliques are shown in Figure~\ref{fig:beyondboss}. The two SIFT based classifiers are significantly worse than BOSS, there is no significant difference between BOSS and SP$_{BD}$, but SP$_{HI}$ is significantly better than BOSS.

\begin{figure}[!ht]
	\centering
    \begin{tabular}{cc}
    \includegraphics[width=5cm,trim={3cm 8cm 3cm 8cm},clip]{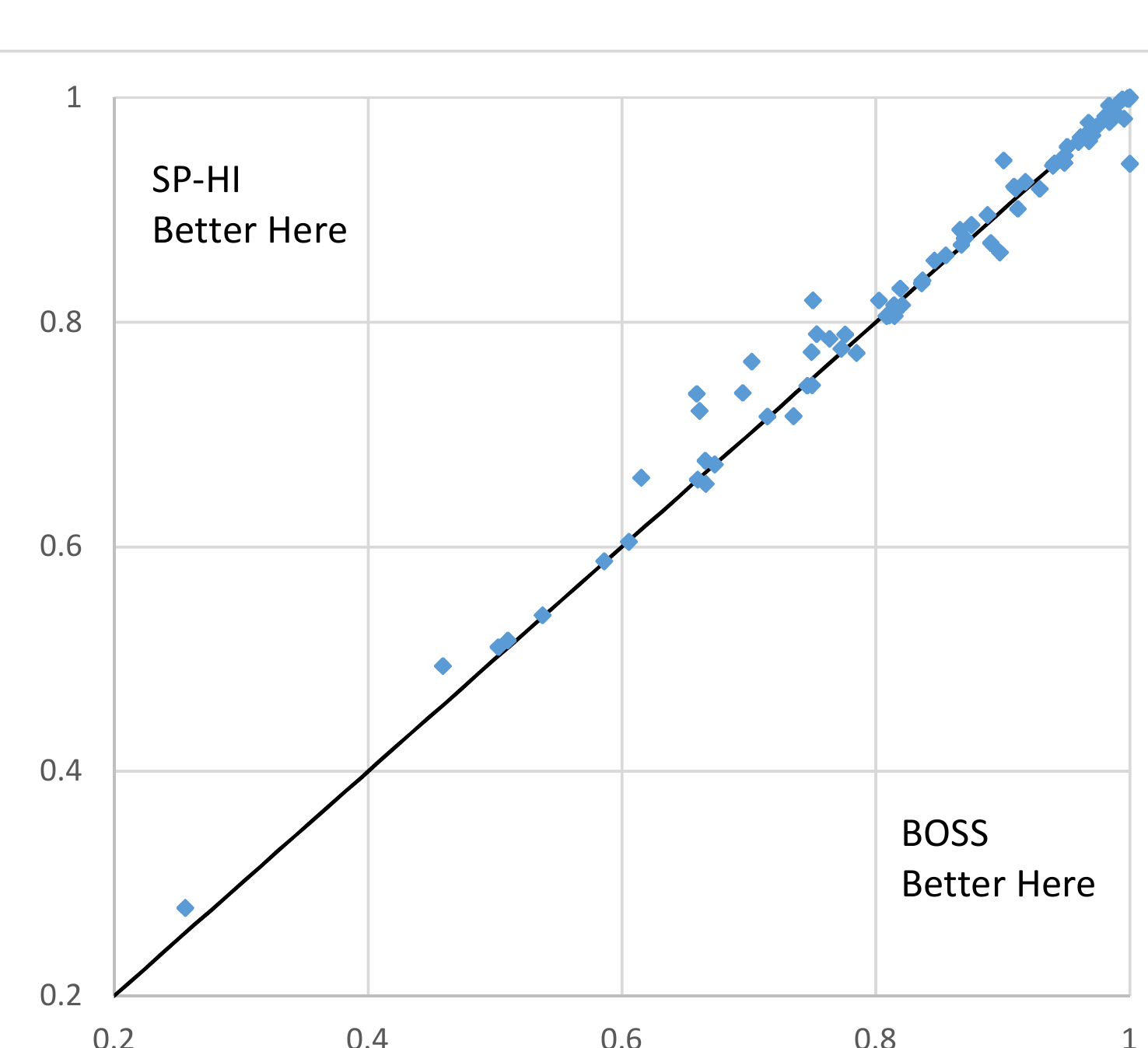} &
        \includegraphics[width=5cm,trim={3cm 8cm 3cm 8cm},clip]{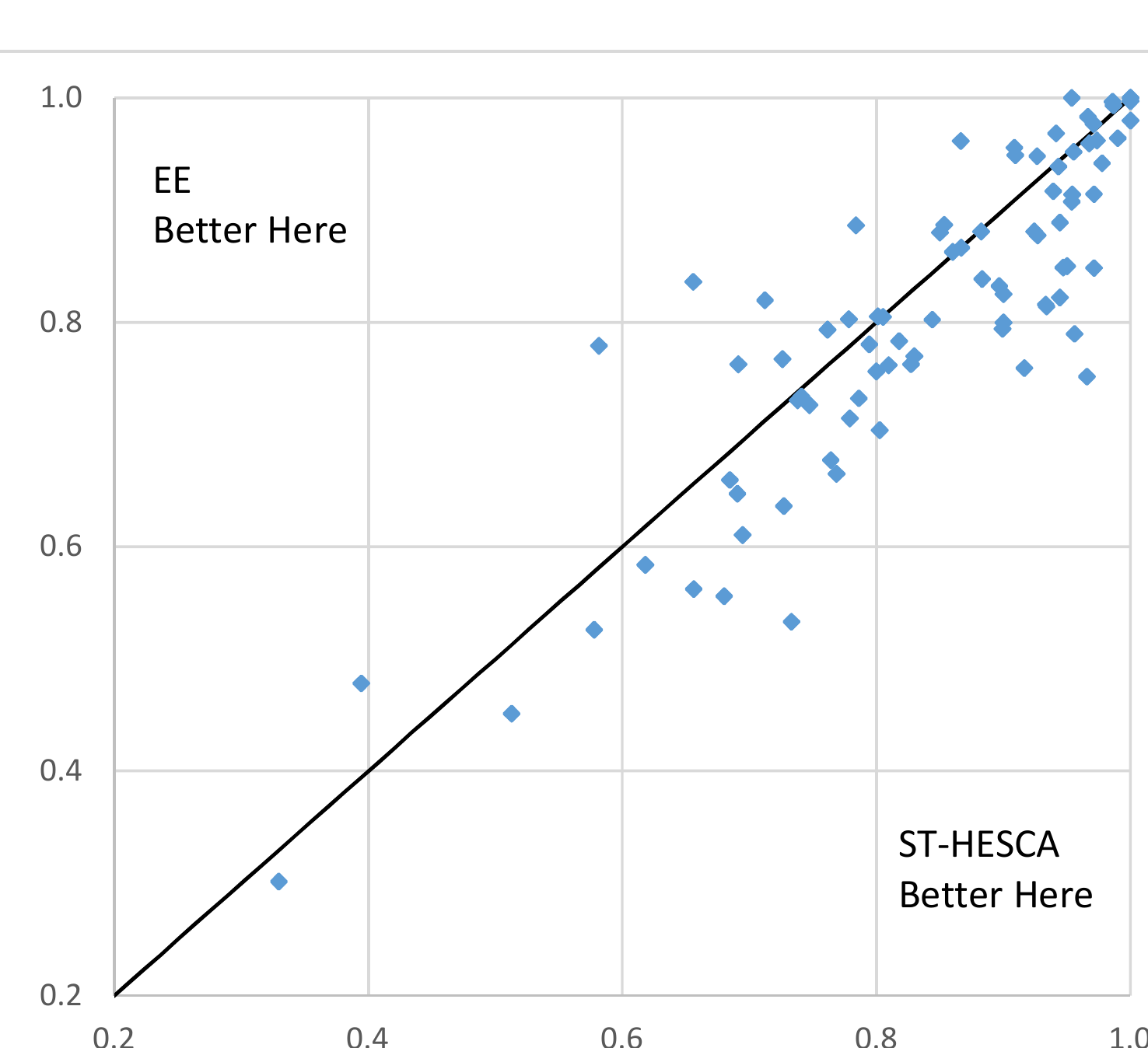}
        \end{tabular}
       \caption{Scatter plot of accuracies for (a) SP$_{HI}$ vs BOSS and (b) EE vs ST-HESCA. The latter is provided to demonstrate the type of spread observable on this data  for two very different classifiers.}
       \label{fig:sp-hi}
\end{figure}

SP$_{HI}$ represents an advance for dictionary based algorithms for TSC, but we do not wish to over sell the importance. Although SP$_{HI}$ wins on 51 of the 77 problems, that actual differences are small. Figure~\ref{fig:sp-hi} shows the scatter plot of accuracies of SP$_{HI}$ against BOSS. The results are fairly tightly grouped on the line of equality. The overall mean difference in accuracy is just 0.64\%. We would expect the improvements from SP$_{HI}$ to be apparent on problems where discriminatory features are phase dependent. For example, SP$_{HI}$ is over 7\% more accurate than BOSS on WordSynonyms and 6\% better on FiftyWords. The elastic ensemble is the most accurate classifier on these two problems, which indicate the discriminatory features are time dependent. Conversely, SP$_{HI}$ is 6\% worse than BOSS on ShapeletSim and 3.5\% worse on SonyAIBORobotSurface1. The phase independent classifier ST-HESCA is the most accurate classifier for these datasets, whereas EE does poorly. This indicates that the setting of the parameter L, like all parameters, is vulnerable to overfitting. Whereas it would have evidently been better to use the regular BOSS classifier (or equivalently setting L=1 in the SP) on those latter datasets due to their phase-independent nature, in terms of training accuracy the parameter search process found some erroneous advantage to using more levels.

\begin{table}
			\centering
			\caption{The mean accuracy of five variants of dictionary classifiers. BOSS~\cite{schafer15boss}, spatial pyramid BOSS with histogram intersection distance (SP-HI) and BOSS distance (SP-BD) and an adapted version of bag of temporal words~\cite{bailly16botsw} with  histogram intersection distance (BOTSW-BD) and BOSS distance (BOTSW-HI)  }
\label{tab:results}
          \scalebox{0.8}{
\begin{tabular}{llllll}
DataSet	 &BOSS	 &SP-HI	 &SP-BD	 &BOTSW-BD	 &BOTSW-HI\\ \hline
Adiac	& {\bf 74.94 ($\pm$0.15) } 	& 74.38 ($\pm$0.25)  	& 74.77 ($\pm$0.26)  	& 71.62 ($\pm$0.4)  	& 71.59 ($\pm$0.29)  \\
ArrowHead	& 87.52 ($\pm$0.33)  	& {\bf 88.66 ($\pm$0.67) } 	& 87.89 ($\pm$0.67)  	& 87.36 ($\pm$0.58)  	& 86.33 ($\pm$0.69)  \\
Beef	& 61.5 ($\pm$0.88)  	& {\bf 66.13 ($\pm$1.49) } 	& 65.2 ($\pm$1.63)  	& 54.93 ($\pm$1.61)  	& 55.6 ($\pm$1.62)  \\
BeetleFly	& {\bf 94.85 ($\pm$0.57) } 	& 94.2 ($\pm$1.31)  	& 93.6 ($\pm$1.28)  	& 91.6 ($\pm$0.95)  	& 92.8 ($\pm$1.04)  \\
BirdChicken	& {\bf 98.4 ($\pm$0.36) } 	& 97.8 ($\pm$1.04)  	& 98 ($\pm$0.87)  	& 95.2 ($\pm$0.84)  	& 92.2 ($\pm$1.39)  \\
Car	& 85.5 ($\pm$0.45)  	& 85.93 ($\pm$0.91)  	& 85.8 ($\pm$1.01)  	& {\bf 90.33 ($\pm$0.77) } 	& 88.2 ($\pm$0.71)  \\
CBF	& 99.81 ($\pm$0.04)  	& {\bf 99.93 ($\pm$0.02) } 	& 99.91 ($\pm$0.03)  	& 99.88 ($\pm$0.05)  	& 99.85 ($\pm$0.07)  \\
ChlorineConcentration	& 65.96 ($\pm$0.13)  	& 65.96 ($\pm$0.29)  	& {\bf 65.97 ($\pm$0.31) } 	& 56.8 ($\pm$4.02)  	& 58.67 ($\pm$11.73)  \\
CinCECGtorso	& 90.05 ($\pm$0.48)  	& {\bf 94.39 ($\pm$0.77) } 	& 93.51 ($\pm$0.87)  	& 78.24 ($\pm$0.95)  	& 71.72 ($\pm$0.81)  \\
Coffee	& {\bf 98.86 ($\pm$0.2) } 	& 98.57 ($\pm$0.36)  	& 98.57 ($\pm$0.36)  	& 98.29 ($\pm$0.51)  	& 98.71 ($\pm$0.41)  \\
Computers	& 80.23 ($\pm$0.23)  	& {\bf 81.92 ($\pm$0.68) } 	& 81.54 ($\pm$0.54)  	& 67.66 ($\pm$0.66)  	& 65.55 ($\pm$0.51)  \\
CricketX	& 76.36 ($\pm$0.2)  	& 78.53 ($\pm$0.42)  	& 77.96 ($\pm$0.4)  	& {\bf 81.22 ($\pm$0.43) } 	& 79.85 ($\pm$0.46)  \\
CricketY	& 74.93 ($\pm$0.18)  	& 77.31 ($\pm$0.32)  	& 76.98 ($\pm$0.41)  	& {\bf 78.61 ($\pm$0.29) } 	& 77.04 ($\pm$0.47)  \\
CricketZ	& 77.57 ($\pm$0.19)  	& 78.88 ($\pm$0.39)  	& 78.51 ($\pm$0.35)  	& {\bf 82.63 ($\pm$0.31) } 	& 82 ($\pm$0.46)  \\
DiatomSizeReduction	& 93.94 ($\pm$0.42)  	& 93.93 ($\pm$0.65)  	& {\bf 94.21 ($\pm$0.68) } 	& 91.19 ($\pm$0.68)  	& 90.99 ($\pm$0.7)  \\
DistalPhalanxOC	& {\bf 81.46 ($\pm$0.21) } 	& 80.52 ($\pm$0.57)  	& 81.25 ($\pm$0.56)  	& 77.61 ($\pm$0.48)  	& 76.09 ($\pm$14.59)  \\
DistalPhalanxOAG	& 81.41 ($\pm$0.28)  	& {\bf 81.5 ($\pm$0.63) } 	& 81.24 ($\pm$0.6)  	& 77.44 ($\pm$0.68)  	& 77.29 ($\pm$0.72)  \\
DistalPhalanxTW	& 67.3 ($\pm$0.21)  	& {\bf 67.34 ($\pm$0.45) } 	& 67.14 ($\pm$0.43)  	& 65.06 ($\pm$0.41)  	& 65.78 ($\pm$0.5)  \\
Earthquakes	& 74.59 ($\pm$0.04)  	& 74.33 ($\pm$0.12)  	& 74.76 ($\pm$0.04)  	& 74.45 ($\pm$0.12)  	& {\bf 74.82 ($\pm$0.15) }\\
ECG200	& {\bf 89.05 ($\pm$0.3) } 	& 87.04 ($\pm$0.64)  	& 87.96 ($\pm$0.63)  	& 86.96 ($\pm$0.55)  	& 86.76 ($\pm$0.59)  \\
ECGFiveDays	& 98.33 ($\pm$0.28)  	& 99.3 ($\pm$0.23)  	& 99.21 ($\pm$0.24)  	& {\bf 99.67 ($\pm$0.23) } 	& 99.33 ($\pm$0.31)  \\
FaceFour	& {\bf 99.56 ($\pm$0.1) } 	& 98.09 ($\pm$0.35)  	& 98.14 ($\pm$0.35)  	& 93.95 ($\pm$0.53)  	& 91.59 ($\pm$0.83)  \\
FacesUCR	& 95.06 ($\pm$0.07)  	& 95.62 ($\pm$0.15)  	& {\bf 95.67 ($\pm$0.15) } 	& 95.31 ($\pm$0.1)  	& 94.38 ($\pm$0.16)  \\
FiftyWords	& 70.22 ($\pm$0.16)  	& 76.5 ($\pm$0.28)  	& {\bf 76.9 ($\pm$0.3) } 	& 75.94 ($\pm$0.38)  	& 72.32 ($\pm$0.24)  \\
Fish	& 96.87 ($\pm$0.11)  	& {\bf 97.1 ($\pm$0.23) } 	& 96.96 ($\pm$0.25)  	& 94.38 ($\pm$0.3)  	& 95.36 ($\pm$0.26)  \\
GunPoint	& 99.41 ($\pm$0.11)  	& {\bf 99.84 ($\pm$0.06) } 	& 99.73 ($\pm$0.1)  	& 98.88 ($\pm$0.13)  	& 98.75 ($\pm$0.1)  \\
Ham	& 83.6 ($\pm$0.38)  	& 83.47 ($\pm$0.85)  	& {\bf 83.62 ($\pm$0.86) } 	& 77.26 ($\pm$1.03)  	& 76.15 ($\pm$1.15)  \\
Haptics	& 45.87 ($\pm$0.36)  	& {\bf 49.36 ($\pm$0.74) } 	& 47.42 ($\pm$0.67)  	& 47.04 ($\pm$0.56)  	& 48.05 ($\pm$0.72)  \\
Herring	& 60.53 ($\pm$0.52)  	& 60.44 ($\pm$1.17)  	& 59.13 ($\pm$1.04)  	& {\bf 60.94 ($\pm$1.32) } 	& 59.94 ($\pm$1.04)  \\
InlineSkate	& 50.26 ($\pm$0.35)  	& 51.08 ($\pm$0.62)  	& {\bf 51.4 ($\pm$0.59) } 	& 42.49 ($\pm$0.44)  	& 40.92 ($\pm$0.65)  \\
InsectWingbeatSound	& 51.03 ($\pm$0.21)  	& 51.67 ($\pm$0.43)  	& {\bf 51.82 ($\pm$0.49) } 	& 50.34 ($\pm$0.27)  	& 46.58 ($\pm$0.31)  \\
ItalyPowerDemand	& 86.6 ($\pm$0.36)  	& 88.22 ($\pm$0.57)  	& 87.14 ($\pm$0.86)  	& {\bf 93.43 ($\pm$0.3) } 	& 92.96 ($\pm$0.31)  \\
LargeKitchenAppliances	& 83.66 ($\pm$0.19)  	& {\bf 83.7 ($\pm$0.6) } 	& 82.02 ($\pm$0.5)  	& 79.16 ($\pm$13.71)  	& 78.24 ($\pm$0.29)  \\
Lightning2	& {\bf 81 ($\pm$0.46) } 	& 80.59 ($\pm$0.97)  	& 80.2 ($\pm$0.93)  	& 79.93 ($\pm$0.86)  	& 79.21 ($\pm$1.04)  \\
Lightning7	& 66.56 ($\pm$0.54)  	& 67.67 ($\pm$1.02)  	& 68.33 ($\pm$1.05)  	& {\bf 73.53 ($\pm$0.78) } 	& 71.07 ($\pm$0.78)  \\
Mallat	& {\bf 94.85 ($\pm$0.12) } 	& 94.8 ($\pm$0.24)  	& 94.8 ($\pm$0.24)  	& 89.35 ($\pm$0.5)  	& 89.44 ($\pm$0.35)  \\
Meat	& 98.03 ($\pm$0.24)  	& {\bf 98.33 ($\pm$0.47) } 	& 98.27 ($\pm$0.48)  	& 95.87 ($\pm$0.67)  	& 96.6 ($\pm$0.61)  \\
MedicalImages	& 71.46 ($\pm$0.23)  	& 71.59 ($\pm$0.32)  	& 72.19 ($\pm$0.33)  	& {\bf 75.02 ($\pm$0.29) } 	& 72.87 ($\pm$0.27)  \\
MiddlePhalanxOC	& {\bf 80.82 ($\pm$0.19) } 	& 80.52 ($\pm$0.4)  	& 80.78 ($\pm$0.36)  	& 75.86 ($\pm$0.4)  	& 75.6 ($\pm$14.8)  \\
MiddlePhalanxOAG	& {\bf 66.6 ($\pm$0.33) } 	& 65.58 ($\pm$0.67)  	& 65.84 ($\pm$0.64)  	& 61.3 ($\pm$0.56)  	& 60.52 ($\pm$0.67)  \\
MiddlePhalanxTW	& 53.74 ($\pm$0.27)  	& {\bf 53.87 ($\pm$0.49) } 	& 53.51 ($\pm$0.55)  	& 53.61 ($\pm$0.43)  	& 53.4 ($\pm$0.47)  \\
MoteStrain	& 84.6 ($\pm$0.31)  	& 85.49 ($\pm$0.72)  	& 85.33 ($\pm$0.62)  	& {\bf 90.06 ($\pm$0.45) } 	& 89.28 ($\pm$0.59)  \\
OliveOil	& 87 ($\pm$0.41)  	& {\bf 87.47 ($\pm$0.78) } 	& {\bf 87.47 ($\pm$0.78) } 	& 86.67 ($\pm$0.72)  	& 86.93 ($\pm$0.94)  \\
OSULeaf	& 96.74 ($\pm$0.1)  	& {\bf 97.79 ($\pm$0.14) } 	& 97.36 ($\pm$0.12)  	& 87.4 ($\pm$0.42)  	& 85.54 ($\pm$0.52)  \\
Phoneme	& 25.62 ($\pm$0.27)  	& {\bf 27.84 ($\pm$0.41) } 	& 27.51 ($\pm$0.47)  	& 21.85 ($\pm$0.34)  	& 18.2 ($\pm$0.31)  \\
Plane	& 99.79 ($\pm$0.04)  	& {\bf 99.89 ($\pm$0.06) } 	& 99.81 ($\pm$0.08)  	& 99.54 ($\pm$0.1)  	& 99.16 ($\pm$0.17)  \\
ProximalPhalanxOC	& 86.74 ($\pm$0.17)  	& {\bf 86.89 ($\pm$0.29) } 	& 86.83 ($\pm$0.32)  	& 79.82 ($\pm$0.37)  	& 79.23 ($\pm$12.11)  \\
ProximalPhalanxOAG	& 81.9 ($\pm$0.22)  	& 83 ($\pm$0.31)  	& 82.85 ($\pm$0.38)  	& {\bf 83 ($\pm$0.44) } 	& 82.79 ($\pm$0.46)  \\
ProximalPhalanxTW	& 77.28 ($\pm$0.22)  	& {\bf 77.62 ($\pm$0.43) } 	& 77.48 ($\pm$0.39)  	& 75.32 ($\pm$0.4)  	& 75.3 ($\pm$0.34)  \\
RefrigerationDevices	& {\bf 78.46 ($\pm$0.37) } 	& 77.26 ($\pm$1.29)  	& 77.28 ($\pm$1.25)  	& 57.33 ($\pm$11.5)  	& 67.34 ($\pm$0.83)  \\
ScreenType	& 58.6 ($\pm$0.3)  	& {\bf 58.69 ($\pm$0.85) } 	& 58.56 ($\pm$0.67)  	& 51.66 ($\pm$0.73)  	& 45.57 ($\pm$0.67)  \\
ShapeletSim	& {\bf 100 ($\pm$0) } 	& 94.09 ($\pm$0.52)  	& {\bf 100 ($\pm$0) } 	& 99.98 ($\pm$0.02)  	& 98.91 ($\pm$0.23)  \\
SmallKitchenAppliances	& 75.02 ($\pm$0.42)  	& {\bf 81.92 ($\pm$0.41) } 	& 77.74 ($\pm$0.71)  	& 67.07 ($\pm$13.14)  	& 62.28 ($\pm$6.74)  \\
SonyAIBORobotSurface1	& {\bf 89.74 ($\pm$0.43) } 	& 86.22 ($\pm$0.92)  	& 89.36 ($\pm$1.12)  	& 89.26 ($\pm$0.74)  	& 86 ($\pm$0.97)  \\
SonyAIBORobotSurface2	& 88.77 ($\pm$0.31)  	& {\bf 89.52 ($\pm$0.51) } 	& 88.04 ($\pm$0.55)  	& 88.21 ($\pm$0.57)  	& 86.06 ($\pm$0.71)  \\
SwedishLeaf	& 91.77 ($\pm$0.09)  	& {\bf 92.51 ($\pm$0.17) } 	& 92.26 ($\pm$0.22)  	& 89.08 ($\pm$14.55)  	& 89.21 ($\pm$13.63)  \\
Symbols	& 96.12 ($\pm$0.15)  	& 96.47 ($\pm$0.21)  	& 96.25 ($\pm$0.23)  	& {\bf 97.06 ($\pm$0.23) } 	& 96.23 ($\pm$0.31)  \\
SyntheticControl	& 96.79 ($\pm$0.09)  	& 96.15 ($\pm$0.21)  	& 96.68 ($\pm$0.16)  	& {\bf 99.53 ($\pm$0.07) } 	& 98.68 ($\pm$0.14)  \\
ToeSegmentation1	& 92.88 ($\pm$0.3)  	& 91.88 ($\pm$0.6)  	& 92.44 ($\pm$0.63)  	& {\bf 93.98 ($\pm$0.34) } 	& 91.58 ($\pm$0.56)  \\
ToeSegmentation2	& 95.97 ($\pm$0.18)  	& 96.03 ($\pm$0.29)  	& 96.15 ($\pm$0.32)  	& {\bf 97.08 ($\pm$0.21) } 	& 95.75 ($\pm$0.32)  \\
Trace	& 99.99 ($\pm$0.01)  	& {\bf 100 ($\pm$0) } 	& {\bf 100 ($\pm$0) } 	& {\bf 100 ($\pm$0) } 	& {\bf 100 ($\pm$0) }\\
TwoLeadECG	& 98.45 ($\pm$0.14)  	& {\bf 98.65 ($\pm$0.24) } 	& 98.54 ($\pm$0.25)  	& 97.23 ($\pm$0.32)  	& 96.6 ($\pm$0.33)  \\
UWaveGestureLibraryY	& 66.12 ($\pm$0.11)  	& {\bf 72.09 ($\pm$0.3) } 	& 71.79 ($\pm$0.26)  	& 71.75 ($\pm$14.35)  	& 69.67 ($\pm$13.34)  \\
Wine	& {\bf 91.17 ($\pm$0.63) } 	& 90.07 ($\pm$1.39)  	& 90.07 ($\pm$1.4)  	& 83.41 ($\pm$1.71)  	& 84.74 ($\pm$1.79)  \\
WordSynonyms	& 65.88 ($\pm$0.19)  	& {\bf 73.62 ($\pm$0.31) } 	& 73.47 ($\pm$0.34)  	& 72.03 ($\pm$0.28)  	& 68.56 ($\pm$0.38)  \\
Worms	& {\bf 73.49 ($\pm$0.45) } 	& 71.64 ($\pm$0.77)  	& 72.42 ($\pm$1.05)  	& 72.16 ($\pm$0.78)  	& 72.31 ($\pm$0.88)  \\
WormsTwoClass	& {\bf 80.97 ($\pm$0.43) } 	& 80.62 ($\pm$0.92)  	& 80.62 ($\pm$0.89)  	& 79.74 ($\pm$0.78)  	& 78.23 ($\pm$0.83)  \\
Yoga	& 90.99 ($\pm$0.13)  	& {\bf 91.89 ($\pm$0.32) } 	& 91.52 ($\pm$0.34)  	& 88.86 ($\pm$3.63)  	& 88.71 ($\pm$0.21)  \\
wins & 16 & 26.75 & 8.25 & 15.25 & 1.25 \\ \hline

\end{tabular}
}
\end{table}

\begin{figure}[!ht]
	\centering
    \includegraphics[width=\linewidth,trim={3cm 14cm 3cm 12cm},clip]{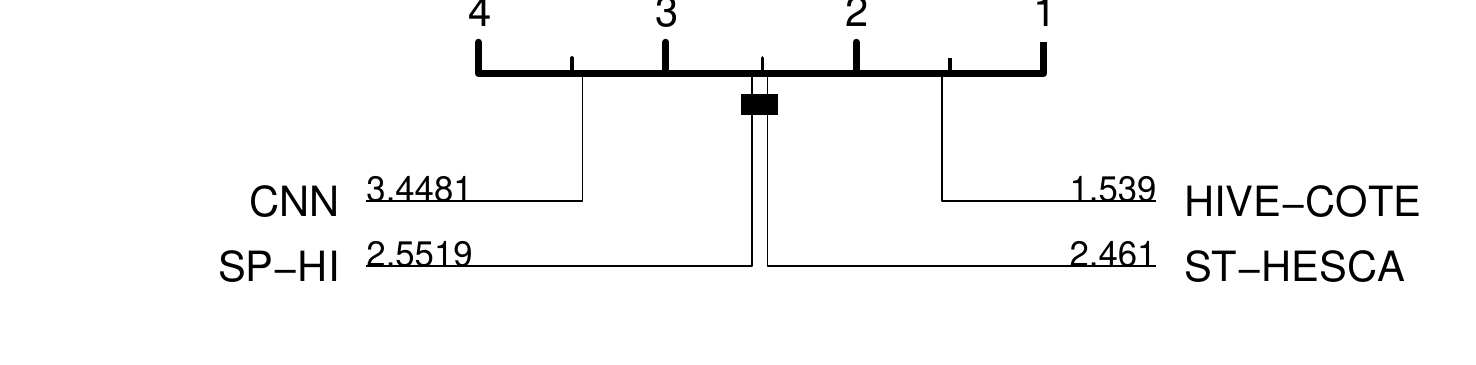}
       \caption{A comparison of SP-HI to three alternative TSC approaches that are not dictionary based.}
       \label{fig:cnn}
\end{figure}
For reference, we compare SP-HI to three state of the art algorithms from alternative domains: a convolutional neural network (CNN); a shapelet transform with a heterogenous ensemble (ST-HESCA); and an ensemble of classifiers from the different transform domains (HIVE-COTE). We compare against ST-HESCA and HIVE-COTE as they have been shown to be significantly better than competing methods~\cite{bagnall17bakeoff}. The details of these classifiers is given in~\cite{lines16hive}. The results, shown in Figure~\ref{fig:cnn} demonstrate that SP-HI is significantly better than CNN and not significantly worse than ST-HESCA. It is significantly worse than HIVE-COTE, but this is to be expected over so many data sets because HIVE-COTE contains BOSS as a component. On average, SP-HI is competitive with state of the art classifiers from different problem domains.

\section{Conclusions}
\label{sec:conc}

Dictionary classifiers are an important class of TSC algorithm that explicitly use the frequency of occurrence of repeating patterns as classification features. A previous study observed a huge difference in accuracy between two prominent approaches, BOP~\cite{lin12bagofpatterns} and BOSS~\cite{schafer15boss}. In order to investigate why this is so, we identify the four key differences between the two algorithms and assess their importance to both BOP and BOSS. We find that only one of these features, ensembling over different parameter values, is beneficial to both BOSS and BOP. Ensembling has proven successful in other domains for TSC~\cite{bagnall15cote}, and it carries no train time overhead if a parameter search is being conducted. BOP with an ensemble is significantly better than a single BOSS classifier. Hence, we would recommend anyone assessing a new TSC algorithm attempt to ensemble, not least to make a better comparison to the state of the art. For example, it is quite possible that HOG-1D+DTW-MDS~\cite{zhao16descriptors} would be significantly more accurate if ensembled.

However, there is more to BOSS than the ensemble. The three other distinguishing features: the use of the Fourier transform, data driven discretisation and bespoke distance measure, all have a significant effect overall. This demonstrates that algorithm design is not always a linear process; algorithm components interact in surprising ways. This is most clearly illustrated with distance measures. The BOSS distance makes BOSS significantly more accurate, but it makes BOP significantly worse. The importance of the distance function is further demonstrated with our experiments involving histogram interaction (HI) distance and two alternative dictionary classifiers, bag of temporal SIFT features (BOTSW) and BOSS with Spatial Pyramid (SP). Using HI made BOTSW significantly worse, but it improved SP (albeit not significantly).

Ensembling does have a memory overhead, as each base classifier must be stored. This is particularly memory intensive for histogram based nearest neighbour classifiers such as BOP and BOSS, and it would be useful to have an algorithm that did not require storing all the histograms in the ensemble. We have experimented with using alternative less memory intensive base classifiers such as C4.5, but this significantly reduced accuracy and massively increased the time to build the classifier. A SVM approach may yield a better classifier with lower memory overhead, but our preliminary experiments showed that the extra training time made this infeasible for a large number of problems. However, it is possible to pursue this further, perhaps through using a condensed data set and/or a proxy classifier for parameter search.


The new approach to dictionary classifiers that combine temporal and dictionary features by using Spatial Pyramids in conjunction with BOSS and HI is significantly more accurate than the standard BOSS ensemble. However, the improvement is small and mostly on problems where BOSS is not the best algorithm, so it is debatable whether the extra memory overhead required by the spatial pyramid is worth the small improvement. We believe the SP approach will be best when discriminatory shape frequency features are embedded in confounding noise. In this situation, the pyramid will facilitate higher pattern resolution in certain areas of the data. Other techniques may also improve classification for certain data, although this has yet to be conclusively shown.

The challenge for dictionary based classifiers is to form a qualitative understanding of the type of problems that best suit this approach and to back this understanding with experimental evidence. For example, we could argue that dictionary classifiers will be a good choice of algorithm for classifying long EEG series. This seems reasonable, given BOSS is based on frequency of repetition patterns, but we have no evidence that this is actually the case. It will then be much easier to quantify whether further possible refinements based on techniques used in other fields actually improve accuracy on data for which it is sensible to use a dictionary classifier.

\section*{Acknowledgements}
This work is supported by the UK Engineering and Physical Sciences Research Council (EPSRC)  [grant number EP/M015087/1]. The experiments were carried out on the High Performance Computing Cluster supported by the Research and Specialist Computing Support service at the University of East Anglia.

\bibliographystyle{plain}
\bibliography{TSCMaster}

\end{document}